\pdfoutput=1

\documentclass[11pt,a4paper]{article}

\usepackage{placeins}
\usepackage{authblk}

\usepackage[hang,flushmargin]{footmisc}
\usepackage[hyperref]{acl}

\usepackage{times}
\usepackage{latexsym}
\usepackage{textcomp}  
\usepackage{bbm}

\usepackage{caption}
\usepackage{float}
\usepackage{placeins}
\usepackage{afterpage}
    
\usepackage[latin1]{inputenc}
\usepackage[T1]{fontenc}
\usepackage{microtype}
\usepackage{url}
\usepackage{booktabs}
\usepackage{tabularx}
\usepackage{ltablex}
\usepackage{xtab}

\usepackage{nicefrac}

\usepackage{algorithm}
\usepackage{algorithmicx}
\usepackage[noend]{algpseudocode}
\algrenewcommand\algorithmicindent{0.8em}

\usepackage{fontawesome}

\newcommand{\Ni}{(1)~}
\newcommand{\Nii}{(2)~}
\newcommand{\Niii}{(3)~}
\newcommand{\Niv}{(4)~}

\usepackage{amsmath}
\usepackage{amssymb} 
\DeclareMathOperator*{\argmax}{argmax}

\usepackage{mathtools}
\DeclarePairedDelimiterX{\kldivx}[2]{(}{)}{%
  #1\;\delimsize\|\;#2%
}

\usepackage{graphicx}
\graphicspath{{figures/}}
\DeclareGraphicsExtensions{.pdf,.jpg,.png}
\setkeys{Gin}{pagebox=artbox}

\usepackage{stfloats}

\usepackage{multirow}
\usepackage{siunitx}
\usepackage{xstring}
\newcommand{\stddev}[1]{%
\IfDecimal{#1}{\hspace{1.5pt}\raisebox{0.5pt}{\scriptsize \num{#1}}}{#1}
}

\newcolumntype{P}{p{0.3cm}}

\usepackage{array}
\newcolumntype{x}[1]{>{\centering\arraybackslash\hspace{0pt}}p{#1}}
\newcolumntype{R}[1]{>{\raggedleft\arraybackslash\hspace{0pt}}p{#1}}
\newcolumntype{L}[1]{>{\raggedright\arraybackslash\hspace{0pt}}p{#1}}

\newcommand{\faNocheck}{{\color{gray}\faRemove}}

\newcommand{\stAcr}{HAST}
\newcommand{\totalRuntime}{2600~hours}

\begin{document}
\title{Self-Training for Sample-Efficient Active Learning for Text Classification with Pre-Trained Language Models}

\date{}

\author[1,2,3]{\textbf{Christopher Schr{\"o}der}}
\author[1,3]{\textbf{Gerhard Heyer}}

\affil[1]{Center for Scalable Data Analytics and Artificial Intelligence (ScaDS.AI), Dresden/Leipzig}
\affil[2]{TUD Dresden University of Technology}
\affil[3]{Leipzig University}

\hypersetup{
  pdftitle={Self-Training for Sample-Efficient Active Learning for Text Classification with Pre-Trained Language Models},
  pdfsubject={cs.CL,cs.LG,cs.AI},
  pdfauthor={Christopher Schr{\"o}der, Gerhard Heyer},
  pdfkeywords={active learning, text classification, self-training, llm}}

\maketitle

\begin{abstract}
Active learning is an iterative labeling process that is used to obtain a small labeled subset, despite the absence of labeled data, thereby enabling to train a model for supervised tasks such as text classification.
While active learning has made considerable progress in recent years due to improvements provided by pre-trained language models, there is untapped potential in the often neglected unlabeled portion of the data, although it is available in considerably larger quantities than the usually small set of labeled data.
In this work, we investigate how self-training, a semi-supervised approach that uses a model to obtain pseudo-labels for unlabeled data, can be used to improve the efficiency of active learning for text classification.
Building on a comprehensive reproduction of four previous self-training approaches, some of which are evaluated for the first time in the context of active learning or natural language processing, we introduce \stAcr{}, a new and effective self-training strategy, which 
is evaluated on four text classification benchmarks. Our results show that it outperforms the reproduced self-training approaches and reaches classification results comparable to previous experiments for three out of four datasets, using as little as 25\% of the data. The code is publicly available at \href{https://github.com/chschroeder/self-training-for-sample-efficient-active-learning}{https://github.com/chschroeder/self-training-for-sample-efficient-active-learning}.
\end{abstract}

\section{Introduction}

In supervised machine learning, a lack of labeled data is the main obstacle to real-world applications, since labeled data is usually non-existent, expensive to obtain, and sometimes even requires domain experts for annotations. 
One solution to create models despite the absence of labels, is {\em{active learning}}, where in an iterative process an oracle (usually realized through a human annotator) provides labels for unlabeled instances that have been deemed to be informative by a so-called {\em query strategy}. 
These labels are then used to train a model, which in turn is used by the query strategy during the next iteration.
In this work, we investigate the combination of self-training and active learning to reduce the required amount of labeled data even further.

\begin{figure}[!t]
\hspace{-2pt}\includegraphics[width=0.484\textwidth]{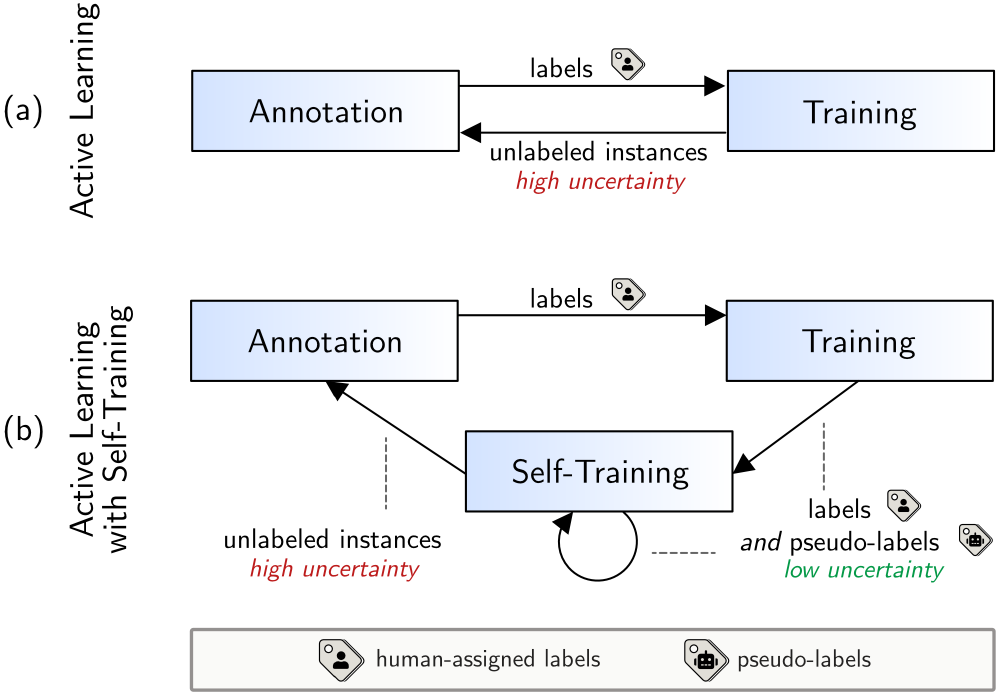}
\caption{
Active learning~(a), and active learning with interleaved self-training~(b).
For active learning, the most uncertain samples are labeled by the human annotator,
while for self-training pseudo-labels are obtained from the current model using the most certain samples.    
}
\label{fig:active-learning-and-self-training}
\end{figure}
During recent years, transformer-based pre-trained language models \citep{vaswani:2017,devlin:2019} have successfully been applied for active-learning-based text classification, thereby considerably raising the state-of-the-art results (e.g., \citeauthor{margatina:2021}, \citeyear{margatina:2021}). 
The dominant paradigm here is pool-based active learning \citep{lewis:1994} where the query strategy repeatedly selects batches of instances to be labeled next from the {\em pool}, the entirety of unlabeled data. 
While language models have successfully been adopted for active learning (e.g., by \citet{ein-dor:2020}, \citet{yuan:2020}, and \citet{margatina:2021}), the total labeling effort, i.e. the number of queries and the number of instances per query, has remained similar to setups predating transformers. 
With regard to the size of queries, there are two prevailing setups: 
\Ni Absolute query sizes \citep{yang:2009,sharma:2015,zhang:2017,ein-dor:2020,yuan:2020,schroeder:2022,tonneau:2022}, where a fixed number of instances are queried during each iteration, and 
\Nii relative query sizes \citep{lowell:2019,prabhu:2019,margatina:2021}, in which the number of queried instances is a percentage of the unlabeled pool. 
We argue that those query sizes of both aforementioned experiment setups are needlessly large. Previous works query up to 1000 instances \citep{yang:2009} or up to 25\% percent of the unlabeled pool \citep{lowell:2019}, where the former is of considerable size and the latter is clearly infeasible in practice as soon as datasets reach average contemporary sizes or annotation costs are high. When using language models that have been trained on billions~\citep{devlin:2019} or even trillions~\citep{touvron:2023} of tokens,
\textbf{there is no need to label hundreds or even thousands of instances.}

In this work, we introduce a sample-efficient active learning approach which incorporates self-training to reduce the amount of training data required for our key task of text classification. Our contributions are as follows:
\Ni~We propose a simple yet highly effective self-training approach that complements high-quality active learning labels with high-quantity {\em pseudo-labels}.
\Nii~We reproduce four existing self-training approaches, enabling a fair comparison among them despite strongly diverging settings and hyperparameter choices in the original works.
\Niii~In extensive experiments, we compare the new approach to the four reproduced methods on four text classification benchmarks using two query strategies and a baseline.
\Niv~ Finally, we discuss possible implications for active learning that result from the observed effectiveness, as well as the trade-offs between favoring small versus large language models. 

The new approach complements active learning with pseudo-labels obtained from the current model.
Using as few as 130 instances, we achieve scores competitive with regard to the state of the art on three out of four datasets.

\section{Related Work}

In this work, we investigate the intersection of self-training and active learning for text classification.

\paragraph{Self-Training} The idea of self-training \citep{scudder:1965, yarowsky:1995} is to leverage unlabeled data in supervised tasks, by obtaining algorithmically-derived pseudo-labels that are subsequently used for training a model.
In natural language processing~(NLP), self-training is an established and well-studied semi-supervised approach \citep{clark:2003,mihalcea:2004,tomanek:2009,ye:2020} that provides additional data by generating soft or hard labels from unlabeled data. 
Similar to active learning, the selection of suitable unlabeled instances is essential. 
However, unlike active learning, there is no human in the loop, therefore, self-training aims to obtain pseudo-labels that are likely to be correct. 
Pseudo-labels, however, are not guaranteed to be correct, a central issue is pseudo-label regularization, which prevents overfitting on incorrect pseudo-labels.

The recently dominant class of pre-trained transformer models \citep{vaswani:2017,devlin:2019} are well-known for their improved effectiveness which derives, among other things, from a high sample efficiency of the contextualized representation, and therefore each additional pseudo-labeled instance can be highly valuable for self-training a pre-trained model.
Consequently, it is unsurpising that self-training has been investigated in NLP with recent model architectures~\citep{meng:2020,mukherjee:2020,vu:2021,gera:2022,chen:2022,sosea:2022}, however, it is still underresearched regarding active learning~\citep{yu:2022,xu:2023}, where the additional labeled data could help to alleviate the data scarcity.

\paragraph{Active Learning and Self-Training}

Despite the recent performance gains achieved by transformer models, active learning still uses a considerable amount of data. 
Recent work in transformer-based active learning for NLP, however, focused on query strategies \citep{ein-dor:2020,margatina:2021,zhang:2021,wertz:2023,zeng:2023}, which often raised the state-of-the-art results given the same labeling budget, but mostly disregarded translating these improvements into reduced annotation efforts.

Despite the comparably slow adoption of self-training, a few recent works already started to investigate the use of unlabeled data in order to improve data efficiency \citep{simeoni:2020,gonsior:2020,gilhuber:2022,tsvigun:2022}. 
In the context of active learning for text classification, both \citet{yu:2022} and \cite{xu:2023} use pre-trained language models for active learning for text classification, thereby outperforming regular active learning.
Their pseudo-label selection, however, relies on the prediction of previous rounds, which renders subsampling, a common method for handling large datasets or expensive models, impossible.
The work of \cite{xu:2023} is closest to our work due to an intersection of self-training and active learning, and text classification.

\section{Active Learning and Self-Training}

The goal for active learning is to minimize the annotation effort while maximizing performance regarding some task, such as text classification.\\

\noindent\textbf{Problem Formulation} In pool-based active learning, the training data $\mathcal{X}=\{(x_i)\}_{i=1}^n$ is partitioned into two disjoint sets: unlabeled pool $\mathcal{U}$ and labeled pool $\mathcal{L}$ (i.e.,~$\mathcal{U} \cap \mathcal{L} = \emptyset$). During the active learning loop, the query strategy selects the best ranked instances $\mathcal{X}_{q} \subseteq \mathcal{U}$, which are then removed from $\mathcal{U}$, labeled by the oracle, and subsequently added to $\mathcal{L}$. 
We refer to a model as $M$, and more specifically as $M_t$ when it has been trained after query $t$.
We denote the predicted class distribution of instance $x$ during query $t$ (using model $M_{t-1}$) as $P_t(y|x)$.

\begin{table*}[!ht]%
\centering
\fontsize{9pt}{10pt}\selectfont%
\renewcommand{\tabcolsep}{5.2pt}%
\begin{tabular}[t]{@{}l@{\hspace{4mm}}x{1cm}x{1cm}x{1cm}ccccccccc@{}}
\toprule
& \multicolumn{3}{c}{\bfseries Pseudo-Label Selection} & \multicolumn{7}{c}{\bfseries Self-Training} & \multicolumn{2}{c}{\bfseries Setting}\\
\cmidrule(r){2-4}\cmidrule(l){5-11}\cmidrule(l){12-13} & \bfseries Subs. & \bfseries Unc. & \bfseries Div. & \bfseries Cls. Bal. & \bfseries Weight. & \multicolumn{5}{c}{\bfseries Regularization} & \bfseries Data  & \bfseries Domain  \\
\cmidrule{7-11} \bfseries Approach & & & & & & D & P & E & T & N & \\
\midrule
UST & \faCheck & \faCheck & (\faCheck) & \faCheck & \faCheck & \faCheck & \faNocheck & \faNocheck & \faNocheck & \faNocheck & Text & Few-Shot\\
AcTune & \faNocheck  & \faCheck & \faCheck & \faNocheck & \faCheck & \faNocheck & \faCheck & \faNocheck & \faCheck & \faNocheck & Text & Active Learning\\
VERIPS & \faCheck &  \faCheck & \faNocheck & \faNocheck & \faNocheck & \faNocheck & \faNocheck & \faCheck & \faCheck & \faNocheck & Images & Active Learning\\
NeST & \faNocheck &  \faCheck & \faNocheck & \faNocheck & \faNocheck & \faNocheck & \faCheck & \faNocheck & \faCheck & \faCheck & Text & Active Learning\\
\midrule
\stAcr{}~(ours) & \faCheck & \faCheck & (\faCheck) & \faCheck & \faCheck & \faNocheck & \faNocheck & \faNocheck & \faCheck & \faCheck & Text & Active Learning\\
\bottomrule
\end{tabular}
\caption{Comparing the four most relevant self-training approaches in terms of pseudo-label selection, self-training, and experiment setting. Symbols: \faCheck:~covered; (\faCheck):~implicitly covered; \faNocheck:~not covered. Abbreviations in the section pseudo-label selection: subsampling (Subs.), uncertainty (Unc.), diversity (Div.). Abbreviations in the section self-training: class balance (Cls. Bal.) and weighting (Weight.). Abbreviations in the section regularization: dropout (D), previous prediction (P), ensembling (E), thresholding (T), and embedding space neighborhood (N).}
\label{table-self-training-approaches}
\end{table*}

\subsection{Incorporating Self-Training}

While active learning aims to obtain a small labeled subset, during training it disregards the data in the unlabeled pool.
Self-training is a semi-supervised approach that, in addition to the labeled pool's data, leverages (parts of) the unlabeled pool by assigning machine-generated pseudo-labels. 
Similar to active learning, it {\em queries} instances according to a criterion such as amongst others, uncertainty. 
In contrast to active learning, however, the selected instances are pseudo-labeled according to some heuristic instead of labeled by a human annotator. 

While for active learning selecting the most uncertain instances has been shown to be effective~\citep{lewis:1994,roy:2001,schroeder:2022}, using the most certain instances has been observed to be most beneficial for self-training across several NLP tasks~(e.g., \citet{mihalcea:2004,tomanek:2009,mukherjee:2020}).
While these approaches obviously contradict, they can complement each other, as shown in Figure~\ref{fig:active-learning-and-self-training}: 
vanilla active learning selects by uncertainty, aiming to find instances that provide the most information to the model, while self-training selects instances by certainty, preferring instances whose pseudo-labels are likely to be correct to increase the set of labeled data.

\subsection{Pseudo-Label Regularization} 
\label{sec:regularization-and-error-propagation}

Pseudo-labels are usually derived from a previous model's predictions.
As a consequence, pseudo-labels are not guaranteed to be correct, which introduces label noise to the self-training process. 
In the case of multiple subsequent self-training iterations, this error can propagate over the iterations, resulting in progressively higher levels of noise~\citep{arazo:2020,yu:2022}. 
Therefore, a key issue for self-training is {\em pseudo-label regularization}, where methods carefully select or weight pseudo-labels to minimize the expected noise.

\subsection{Previous Approaches}
\label{sec:previous-approaches}

Similar to active learning, at the heart of each self-training approach is a strategy that decides which instances are selected---but in this case to be pseudo-labeled. In the following, we present the four most relevant self-training approaches.

\paragraph{UST} Uncertainty-aware self-training (UST; \citet{mukherjee:2020}) uses dropout-based stochastic sampling to obtain multiple confidence estimates for each instance.
Aggregated scores are then obtained using the BALD measure~\citep{houlsby:2011}, based on which instances are sampled in a class-balanced manner.
 
\paragraph{AcTune} AcTune~\citep{yu:2022} aims to obtain a diverse set of instances by preceding the sampling step with weighted K-Means clustering. 
To overcome the noise of per-instance label variation during self-training iterations, it aggregates pseudo-labels over multiple iterations.

\paragraph{VERIPS} The verified pseudo-label selection (VERIPS; \cite{gilhuber:2022}) starts by selecting instances whose prediction confidence exceeds a fixed threshold.
Pseudo-labels are then filtered by a verification step, retaining only the labels for which the current model's predictions match those of a model trained without pseudo-labels.

\paragraph{NeST} Neighborhood-regularized self-training (NeST; \cite{xu:2023}) leverages the embedding space to obtain pseudo-labels that closely match the predicted distribution of their $k$-nearest neighbors. The individual scores are averaged over multiple active learning iterations for additional stability.\\

\noindent In Table~\ref{table-self-training-approaches}, we compare the distinguishing features of all presented approaches, including the  approach proposed in Section~\ref{sec:new-self-training-method}. 
A striking common feature is that all sample selection mechanisms rely on uncertainty, which has been shown to be very effective both for active learning and self-training~\citep{yu:2022,xu:2023}.
The main difference is the pseudo-label selection and regularization.

\subsection{Limitations of Previous Approaches}
\label{sec:limitations-of-previous-approaches}

Apart from methodological similarities and differences, we also identified several conceptual shortcomings shared by several approaches, which limit the conclusiveness of existing evaluations.

\paragraph{Unrealistic Evaluation Settings} The experiments of UST use validation sets matching the size of the training data, and those of AcTune select the best model based on validation sets of sizes $500$ and $1000$. 
Validation sets of these sizes are unrealistic for an active learning scenario, where even training sets of these sizes would exceed the amount of data that we deem to be necessary when evaluating on common text classification benchmarks. 
Moreover, the classification performance is also supported by an extensive hyperparameter optimization~\citep{yu:2022,xu:2023}, which would not be possible without these validation sets.

\paragraph{Computational Efficiency} Since transformer models are known to be computationally expensive, UST and VERIPS incorporate a subsampling mechanism before pseudo-label selection, thereby enabling the use of self-training even with computationally expensive models and large datasets.
The pseudo-label selection of both AcTune and NeST, however, relies on predictions from previous self-training iterations.
If the current subsample differs from previous ones, these previous predictions will not be available, rendering subsampling impractical for those methods and considerably constraining their computational efficiency.

\paragraph{Confidence Thresholds} VERIPS, AcTune, and NeST apply a strict confidence threshold as part of their pseudo-label regularization, which has been shown to have considerable impact on the performance~\citep{gilhuber:2022,yu:2022}.
VERIPS and NeST keep it fixed at a high value, while AcTune performs a hyperparameter search for each dataset. 
The former relies on the availability of high confidence instances, and the latter is infeasible in real-world active learning scenarios, where no validation data exists.
\\

\noindent The limitations illustrated above cast doubt on the generalizability of these findings. 
Furthermore, differences in task (text and image classification) and setting (few-shot and active learning), complicate cross-study comparisons. 
Therefore, active learning and self-training require further investigation, which we address through a reproduction study.

\section{Hard-Label Regularized Self-Training}
\label{sec:new-self-training-method}

Based on a methodological analysis (Sections~\ref{sec:previous-approaches} and~\ref{sec:limitations-of-previous-approaches}) and a reproduction study (Section~\ref{sec:experiments}), we present \textbf{ha}rd-label neighborhood-regularized \textbf{s}elf-\textbf{t}raining (HAST, pronounced {\em ``haste''}), a novel self-training method that aims to complement active learning with large quantities of pseudo-labels.

The idea of our proposed approach is to rely on the generalization capabilities provided by {\em contrastive (representation) learning} and on the regularization provided by nearest neighbor relationships in the resulting embedding space. In contrastive learning, training is performed with $n$-tuples of instances, in the following assumed to be pairs.
A pair can be formed between any two instances, and consequently each additional pseudo-labeled instance increases the number of possible pairings.
Obviously, when combining this with self-training, this can considerably enhance the effective number of instances---at the risk of introducing noise due to incorrect pseudo-labels.

\subsection{Contrastive Representation Learning} Commonly used representations that are obtained from a language model's layers, such as the \texttt{[cls]} token~\citep{devlin:2019}, rely on the principle that semantically similar inputs will result in similar embedding vectors---but apart from testing for semantic similarity, distance metrics between vectors are often meaningless. 
Representation learning~\citep{bengio:2013} on the other hand, aims to learn a meaningful space in which the dimensions capture explanatory factors in the data~\citep{bengio:2013,le-khac:2020}
and distance metrics are rendered meaningful~\citep{le-khac:2020}.
Moreover, in {\em contrastive} representation learning~\citep{carreira-perpinan:2005}, this is achieved by training on contrasting pairs of instances, where similar instances are pulled together and dissimilar instances are pushed apart in the embedding space.
One recent approach is the fine-tuning paradigm SetFit~\citep{tunstall:2022}, which uses a Siamese network to train embeddings that are then used as representations in downstream tasks.
SetFit has shown incredible effectiveness in the few-shot setting~\citep{tunstall:2022}, making it an obvious choice for active learning.

\begin{algorithm}
\fontsize{10pt}{11pt}\selectfont%
\caption{\textsc{AL with self-training}}
\label{alg:active-learning-loop}
\begin{flushleft}
\textbf{Input:} unlabeled pool $\mathcal{U}$; labeled pool $\mathcal{L}$; initial model $M_0$; number~of queries $Q$; batch size $B$; self-training~iterations~$T$\\
\end{flushleft}
\begin{algorithmic}[1]
    \For{q $\in$ \{1, ..., $Q$\}}
    \State $\mathcal{X}_q \gets \text{query batch of size $B$ from }\mathcal{U}$ \label{active-learning-loop:query}
    \State $\mathcal{Y}_q \gets \text{labels provided by oracle}$
    \State $\mathcal{L} \gets \mathcal{L}  \cup \{(x_{q,i}, y_{q,i}), ..., (x_{q,i+B}, y_{q,i+B})\}$
    \State $\mathcal{U} \gets \mathcal{U}  \setminus \{(x_{q,i}, y_{q,i}), ..., (x_{q,i+B}, y_{q,i+B})\}$
    \State $M_q \gets \textsc{Train}(\mathcal{L})$ \label{active-learning-loop:training}
    \State $M_q^{*} \gets \textsc{SelfTrain}(\mathcal{U}, \mathcal{L}, M_q, $T$)$ \label{active-learning-loop:line-self-training}
    \EndFor
\end{algorithmic}
\begin{flushleft}
\textbf{Output:} Final model $M_Q^{*}$
\end{flushleft}
\end{algorithm}

\subsection{Active Learning and Self-Training}

We incorporate self-training into pool-based active learning by adding a subsequent self-training step after each training step as shown in Algorithm~\ref{alg:active-learning-loop}. After each query (line~\ref{active-learning-loop:query}), a new model is trained~(line~\ref{active-learning-loop:training}), expressed through a generic \textsc{\scshape Train()} function that takes a list of labeled instances, and optionally a list of weights. This is followed by a self-training step (line~\ref{active-learning-loop:line-self-training}), which may be HAST or one of the previous approaches. 
The training in line~\ref{active-learning-loop:training} could be skipped after the first iteration, but this additional step ``resets'' a potentially degraded model, thereby counteracting model instability~\citep{mosbach:2021} and model collapse due to error propagation.

\begin{algorithm}
\fontsize{10pt}{11pt}\selectfont%
\caption{\textsc{SelfTrain (\stAcr)}}
\label{alg:self-training}
\begin{flushleft}
\textbf{Input:} unlabeled pool $\mathcal{U}$; labeled pool $\mathcal{L}$; current Model $M_{t_0}$, number of self-training iterations $T$\\
\end{flushleft}
\begin{algorithmic}[1]
    \State{$\mathcal{L}_p = \mathcal{L}$; $\mathcal{U}_p = \mathcal{U}$}
    \For{t $\in$ \{1, ..., $T$\}}
    \State $\mathcal{Y}_{q,t} \gets M_{t_0}(\mathcal{U}_p)$
    \State $\mathcal{X}_{q,t}^{*} \gets \{x_i | x_i \in \mathcal{U}_p \text{ and } \mathbbm{1}_{PL}(x)\}$ \label{self-learning-loop:query}
    \State $\mathcal{L}_p \gets \mathcal{L}_p  \cup \{(x_{t,i}, y_{t,i}), ..., (x_{t,m}, y_{t,m})\}$
    \State $\mathcal{U}_p \gets \mathcal{U}_p  \setminus \{(x_{t,i}, y_{t,i}), ..., (x_{t,m}, y_{t,m})\}$
    \State $W_{q,t} \gets \text{Weights as described by Eq.~\ref{eq:weights}}$.
    \State $M_{q,t}^{*} \gets \textsc{Train}(\mathcal{L}_p, W_{q,t})$
    \EndFor
\end{algorithmic}
\begin{flushleft}
\textbf{Output:} Self-trained model $M_{q,t}^{*}$
\end{flushleft}
\end{algorithm}

\subsection{HAST: Pseudo-Labels and Weighting}
\label{sec:hard-label-regularization}

\noindent The proposed approach is intended to exploit the current model to ideally provide larger amounts of pseudo-labels by leveraging the embedding space.
Instead of relying on label distributions, we use hard labels, which are obtained by a majority vote of the instance's $k$ nearest neighbors (KNN). The proposed approach is shown in Algorithm~\ref{alg:self-training}. 
Our pseudo-label selection~(line~\ref{self-learning-loop:query}) takes all instances from the unlabeled pool $\mathcal{U}_p$, where the most confident label crosses the binary decision threshold of $0.5$ and the predicted label $\hat{y}_i$ agrees with the k nearest neighbors' majority vote:

\vspace{-\baselineskip}
\begin{equation}
\mathbbm{1}_{PL}(x) = 
\begin{cases}
    1 & \text { if } s_i > 0.5 \wedge \hat{y}_i^{knn} = \hat{y}_i\\
    0 & \text{otherwise}
    \end{cases}
\end{equation}

\begingroup
\thickmuskip=2mu
\noindent where $s_i=P(y_i=\hat{y_i}|x) \in (0, 1]$ is the confidence of the most confident predicted label $\hat{y}_i$, and $\hat{y}^{knn}$ is the label given by a KNN majority vote. Since the predicted label $\hat{y}_i = \argmax P(y|x)$ is obtained from the class with highest confidence, this strategy implicitly selects instances with high certainty. 
\paragraph{Weighting} With the proposed pseudo-label selection strategy, we can obtain a potentially large number of pseudo-labels. This can introduce both \Ni~a~class imbalance~\citep{henning:2023} among the pseudo-labels and \Nii~an imbalance between the pseudo-labels and human-annotated labels.
\endgroup

To overcome these issues, we first introduce a weighting term to adjust for class imbalance:

\begin{equation}
z = \frac{N/C - h_c}{max(1, h_c)}\\
\end{equation}

\noindent where $N$ is the number of all pseudo-labels in $\mathcal{Y}_{q,t}$, $C$ is the number of classes, $N/C$ is the
expected number of instances for a balanced class distribution, and $h_c$ is the count of class $c$ in the histogram of $\mathcal{Y}_{q,t}$ over $c$ bins. A $max$ operator in the denominator makes the function well-defined for $h_c \in \mathbb{N}_0$.

This yields a term that is inversely proportional to the current class imbalance. Since the resulting values are unbounded and can potentially grow very large, we apply a sigmoid function to squash the values into the interval $(0, 10)$:

\begin{equation}
\alpha_{c} = \frac{10}{1 + e^{-z}}
\end{equation}

\noindent To reduce the effect of a possibly excessive number of pseudo-labels and retain some weight on the human-annotated labels, we introduce another term $\beta \in (0, 1]$, which is the labeled-to-unlabeled ratio weight that penalizes pseudo-label weights iff ${\beta<1}$. The final weights are then given by:

\begin{equation}
W_i = \alpha_{\hat{y_i}} \cdot \beta
\label{eq:weights}
\end{equation}

\begingroup
\thickmuskip=2mu
\noindent Human-annotated instances have a weight of $W_i = 1.0$. Finally, the resulting weights are \mbox{$L^{1}$-normalized}, i.e.~$\sum_{i} |W_i|=1$, and are then element-wise multiplied with the per-instance loss.
\endgroup

\section{Experiments}
\label{sec:experiments}

\begin{table}[b!]%
\centering
\fontsize{9pt}{10pt}\selectfont%
\renewcommand{\tabcolsep}{1.8pt}%
\begin{tabular}[t]{@{}l@{\hspace{3pt}}cccrrrrc@{}}
\toprule
\bfseries Dataset Name {\tiny (ID)} & \bfseries Type & \bfseries Classes & \bfseries Training & \bfseries Test & \bfseries Metric\\
\midrule
AG's News {\scriptsize (AGN)}       & N & 4 & 120,000 & 7,600 & Acc.\\
DBPedia-14 {\scriptsize (DBP)} & T   & 14 &  560,000 & 70,000                        & F$_1$\\
IMDB {\scriptsize (IMDB)}    & S   & 2 &  25,000 &                       25,000 & Acc.\\
TREC-6 {\scriptsize (TREC-6)}       & Q   & 6 &   5,500 & 500 & F$_1$ \\
\bottomrule
\end{tabular}
\caption{Key information about the examined datasets. Abbreviations: N (News), S (Sentiment), Q (Questions). 
}
\label{table-datasets}
\end{table}

In the experiments, we evaluate the proposed self-training method \stAcr{}. Moreover, in an extensive reproduction study, we re-implement the four most relevant previous self-training approaches, evaluating them on active learning for text classification.

\subsection{Experiment Setup}

The key task in this work is active learning for single-label text classification.
Using only 130 instances, the experiments are designed to be both challenging and data efficient. 
We compare \stAcr{} against the reproductions of four previous approaches (UST, AcTune, NeST, and VERIPS) which are compared under equivalent conditions.

\paragraph{Data} We evaluate on four established text classification benchmarks, whose key characteristics are displayed in Table~\ref{table-datasets}. AGN and IMDB exhibit a balanced class distribution, and DBP and TREC-6 exhibit an imbalanced class distribution. IMDB is a binary classification problem, while the other datasets are multi-class problems. 

\paragraph{Evaluation} Following~\citet{kirk:2022}, we report the classification performance in accuracy for balanced and in macro-F$_1$ for imbalanced datasets.

\begin{table*}[!ht]%
\centering%
\fontsize{9pt}{10pt}\selectfont%
\renewcommand{\tabcolsep}{10pt}%
\begin{tabular}[t]{@{}l@{\hspace{6mm}}l@{\hspace{6mm}}rrrrr@{}}
  \toprule
  & & & \multicolumn{4}{@{}c@{}}{\bfseries Datasets}\\
  \cmidrule{4-7}\bfseries Query Strategy & \bfseries Classifier & \bfseries Self-Training & \bfseries AGN & \bfseries DBP & \bfseries IMDB & \bfseries TREC\\\midrule
  \multirow{12}{*}{Breaking Ties} & \multirow{6}{*}{BERT} & No Self-Training & 0.763\stddev{0.057} & 0.619\stddev{0.129} & 0.745\stddev{0.030} & 0.341\stddev{0.130}\\
   & & UST & 0.798\stddev{0.016} & 0.645\stddev{0.042} & 0.764\stddev{0.100} & 0.333\stddev{0.121}\\
   & & AcTune & 0.806\stddev{0.021} & 0.651\stddev{0.054} & 0.795\stddev{0.050} & 0.434\stddev{0.063}\\
   & & VERIPS & 0.834\stddev{0.012} & 0.907\stddev{0.047} & 0.816\stddev{0.050} & 0.540\stddev{0.110}\\
   & & NeST & \bfseries 0.840\stddev{0.013} & \bfseries 0.918\stddev{0.006} & 0.783\stddev{0.041} & \bfseries 0.580\stddev{0.090}\\
   & & HAST & 0.762\stddev{0.055} & 0.605\stddev{0.071} & \bfseries 0.806\stddev{0.097} & 0.424\stddev{0.193}\\
   \cmidrule(){3-7} & \multirow{6}{*}{SetFit} & No Self-Training & 0.853\stddev{0.011} & 0.973\stddev{0.004} & 0.872\stddev{0.009} & 0.691\stddev{0.029}\\
   & & UST & 0.658\stddev{0.030} & 0.483\stddev{0.033} & 0.851\stddev{0.028} & 0.491\stddev{0.042}\\
   & & AcTune & 0.863\stddev{0.006} & 0.980\stddev{0.003} & 0.896\stddev{0.024} & 0.642\stddev{0.030}\\
   & & VERIPS & 0.859\stddev{0.005} & 0.981\stddev{0.002} & 0.857\stddev{0.024} & 0.730\stddev{0.021}\\
   & & NeST & 0.878\stddev{0.005} & 0.981\stddev{0.001} & \bfseries 0.927\stddev{0.005} & \bfseries  0.781\stddev{0.034}\\
   & & HAST & \bfseries 0.886\stddev{0.007} & \bfseries 0.984\stddev{0.001} & 0.882\stddev{0.040} & 0.773\stddev{0.024}\\
  \midrule
  \multirow{12}{*}{Contrastive Predictions} & \multirow{6}{*}{BERT} & No Self-Training & 0.635\stddev{0.087} & 0.366\stddev{0.034} & 0.670\stddev{0.065} & 0.210\stddev{0.089}\\
  & & UST & 0.712\stddev{0.153} & 0.203\stddev{0.108} & 0.765\stddev{0.056} & 0.311\stddev{0.108}\\
  & & AcTune & 0.678\stddev{0.178} & 0.353\stddev{0.165} & 0.684\stddev{0.079} & 0.136\stddev{0.076}\\
  & & Verips & \bfseries 0.823\stddev{0.018} & \bfseries 0.655\stddev{0.093} & 0.750\stddev{0.067} & 0.449\stddev{0.108}\\
  & & NeST & 0.701\stddev{0.085} & 0.606\stddev{0.083} & 0.772\stddev{0.079} & \bfseries 0.517\stddev{0.166}\\
  & & HAST & 0.718\stddev{0.071} & 0.327\stddev{0.122} & \bfseries 0.810\stddev{0.045} & 0.288\stddev{0.091}\\
  \cmidrule(){3-7} & \multirow{6}{*}{SetFit} & No Self-Training & 0.798\stddev{0.003} & 0.678\stddev{0.055} & 0.890\stddev{0.016} & 0.560\stddev{0.048}\\
  & & UST & 0.597\stddev{0.025} & 0.400\stddev{0.028} & 0.842\stddev{0.015} & 0.420\stddev{0.069}\\
  & & AcTune & 0.811\stddev{0.012} & 0.704\stddev{0.052} & 0.903\stddev{0.018} & 0.622\stddev{0.061}\\
  & & Verips & 0.814\stddev{0.011} & 0.776\stddev{0.056} & 0.893\stddev{0.013} & 0.638\stddev{0.059}\\
  & & NeST & 0.842\stddev{0.006} & 0.786\stddev{0.094} & \bfseries 0.919\stddev{0.005} & 0.605\stddev{0.055}\\
  & & HAST & \bfseries 0.849\stddev{0.013} & \bfseries 0.815\stddev{0.087} & 0.916\stddev{0.009} & \bfseries 0.773\stddev{0.016}\\
  \midrule
  \multirow{12}{*}{Random} & \multirow{6}{*}{BERT} & No Self-Training & 0.760\stddev{0.040} & 0.534\stddev{0.061} & 0.740\stddev{0.046} & 0.276\stddev{0.100}\\
   & & UST & 0.797\stddev{0.039} & 0.693\stddev{0.083} & 0.794\stddev{0.014} & 0.298\stddev{0.065}\\
   & & AcTune & 0.791\stddev{0.050} & 0.559\stddev{0.082} & 0.801\stddev{0.045} & 0.386\stddev{0.123}\\
   & & VERIPS & 0.812\stddev{0.023} & 0.850\stddev{0.063} & 0.813\stddev{0.029} & 0.551\stddev{0.074}\\
   & & NeST & \bfseries 0.819\stddev{0.039} & \bfseries 0.865\stddev{0.037} & 0.782\stddev{0.022} & \bfseries 0.553\stddev{0.071}\\
   & & HAST & 0.677\stddev{0.114} & 0.650\stddev{0.053} & \bfseries 0.831\stddev{0.051} & 0.514\stddev{0.169}\\
   \cmidrule(){3-7} & \multirow{6}{*}{SetFit} & No Self-Training & 0.848\stddev{0.005} & 0.939\stddev{0.031} & 0.907\stddev{0.007} & 0.676\stddev{0.031}\\
   & & UST & 0.659\stddev{0.038} & 0.476\stddev{0.039} & 0.871\stddev{0.031} & 0.491\stddev{0.061}\\
   & & AcTune & 0.847\stddev{0.014} & 0.970\stddev{0.008} & 0.918\stddev{0.004} & 0.651\stddev{0.025}\\
   & & VERIPS & 0.854\stddev{0.010} & 0.968\stddev{0.008} & 0.921\stddev{0.002} & 0.726\stddev{0.017}\\
   & & NeST & 0.860\stddev{0.011} & 0.965\stddev{0.004} & 0.923\stddev{0.008} & \bfseries  0.797\stddev{0.024}\\
   & & HAST & \bfseries 0.885\stddev{0.002} & \bfseries 0.974\stddev{0.006} & \bfseries 0.926\stddev{0.004} & 0.738\stddev{0.020}\\
  \bottomrule
  \end{tabular}  
  \caption{Classification performance after the final iteration (in accuracy or macro-F$_1$), broken down per query strategy, classifier, and self-training approach. The reported numbers represent the average over five runs, with the standard deviations shown to the right of each value.}
\label{table-self-training-results-acc}
\end{table*}

\paragraph{Classification} 

We evaluate two different models: \Ni the paraphrase-mpnet-base SBERT~\citep{reimers:2019} model, which is fine-tuned using SetFit~\citep{tunstall:2022}, and, in order to verify its effectiveness in the non-contrastive setting, \Nii a BERT-base model~\citep{devlin:2019} that is trained using vanilla fine-tuning. Both of these models consist of 110M trainable parameters.

\paragraph{Active Learning} Models are initialized with 30~instances. Active learning is performed over 10~iterations during each of which 10 more instances are labeled. Following \cite{hu:2018} and \cite{yu:2022}, the model is trained from scratch after each active learning and self-training iteration. While we do not directly investigate query strategies in this work, 
they are paramount to active learning and need to be considered. 
To assess their effect on the self-training process, we evaluate all configurations using two query strategies and a baseline: breaking ties~\citep{scheffer:2001,luo:2005},  contrastive predictions\footnote{This query strategy is originally called {\em contrastive active learning}, referring to contrastive differences between predictive distributions of embedding similar instances. In this work, we refer to this strategy as {\em contrastive predictions} to avoid confusion with contrastive representation learning.}~\cite{margatina:2021}, and random sampling.
\begingroup
\thickmuskip=2mu
\paragraph{Self-Training} For \stAcr{}, we use $k=5$ and ${\beta=0.1}$. For all other strategies, we use the best hyperparameters as reported in the respective publications.
A subsample of $16384$ instances is drawn before obtaining pseudo-labels for all strategies supporting subsampling.
To minimize the effect of error propagation (as shown in Section~\ref{sec:regularization-and-error-propagation}), but also for reasons of computational feasibility, we refrain from consecutive self-training iterations.
\endgroup

\subsection{Results}

The final classification performance of each configuration is shown in Table~\ref{table-self-training-results-acc}.
We observe self-training to be highly effective, with improvements of up to 29 percentage points compared to active learning without self-training. 
The best results (in bold) are always achieved by either NeST or \stAcr{}.
The former wins for most BERT configurations, while the latter wins for most SetFit configurations. 

\begin{figure*}[!t]
\includegraphics[width=\textwidth]{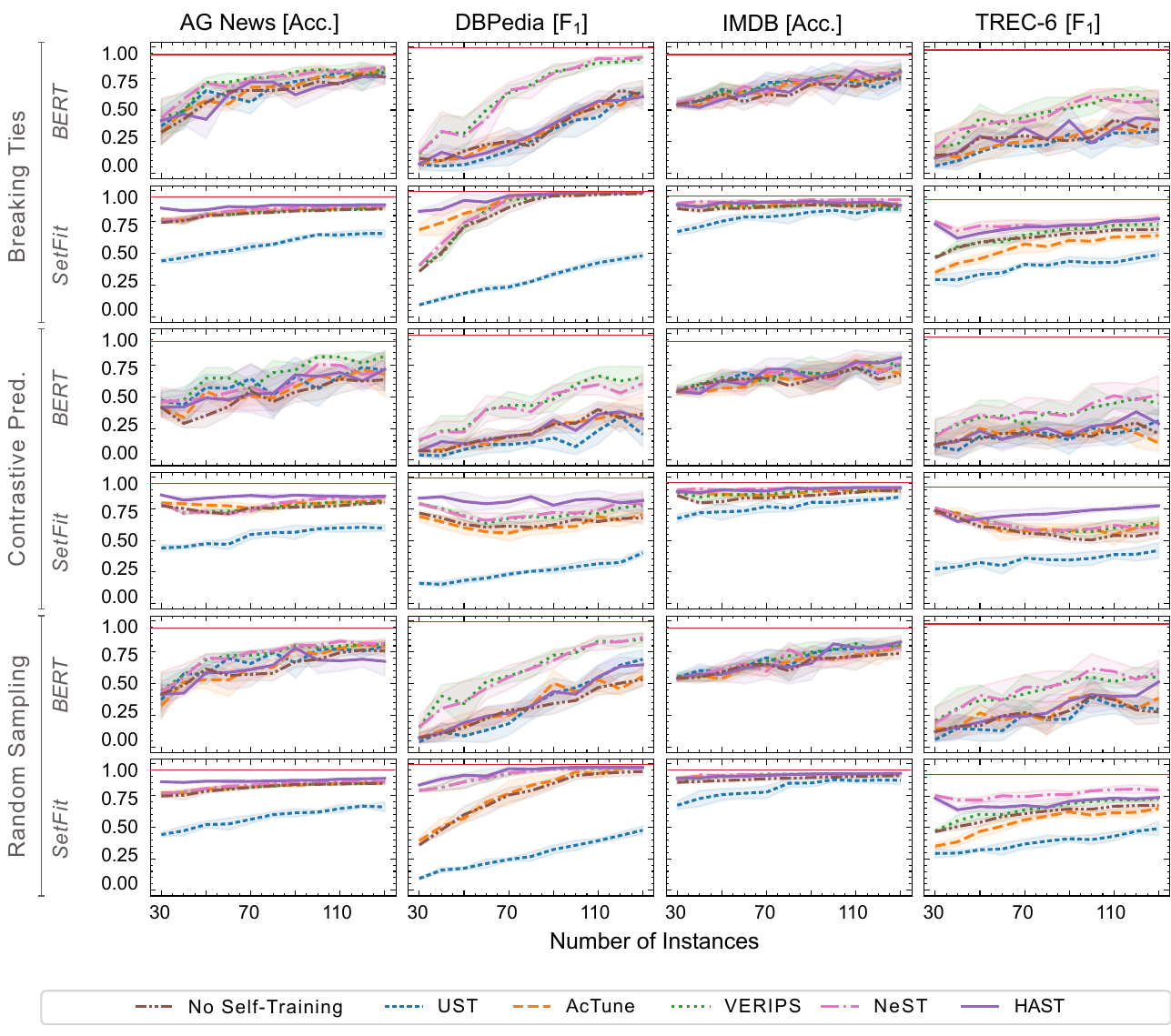}
\caption{Learning curves per model, query strategy, and dataset, showing the classification performance on the test set. The x-axis shows the number of instances, while the y-axis indicates classification performance. The horizontal (red) line represents the performance of the respective model trained on 100\% of the data (without active learning).}
\label{fig:learning-curves}
\end{figure*}
\begin{table}[hbp!]%
\centering
\fontsize{9pt}{10pt}\selectfont%
\renewcommand{\tabcolsep}{9.5pt}%
\begin{tabular}[t]{@{}l@{\hspace{7pt}}llr@{}}
\toprule
\bfseries Dataset & \bfseries Approach (Parameters) & \bfseries N & \bfseries Score \\
\midrule
AGN & ReGen$^1$ (125M)& 0 & 0.850 \\
\small$\left[{}\text{Acc.}\right]$ & BERT$^3$ (336M) & 525 & 0.904\\ 
 & \stAcr{} (110M) & 130 & 0.886\\
 \midrule
DBP & DeBERTa$^4$ (355M)& 0 & 0.945\\
\small$\left[{}\text{F}_1\right]$ & UST$^2$ (110M) & 420 & 0.986\\
& \stAcr{} (110M) & 130 & 0.984\\
\midrule
IMDB & RoBERTa (355M)$^4$& 0 & 0.925\\
\small$\left[{}\text{Acc.}\right]$   & UST$^2$ (110M) & 60 & 0.900 \\
& \stAcr{} (110M) & 130 & 0.927\\
\midrule
TREC-6 & GPT3.5 Turbo \& RoBERTa$^5$ & 0 & 0.914\\
\small$\left[{}\text{F}_1\right]$ & BERT$^3$ (336M) & 525 & 0.968\\
& \stAcr{} (110M) & 130 & 0.773\\

\bottomrule
\end{tabular}
\caption{Comparison with previous works that have investigated low-resource methods: $^1$\citep{yu:2023}, $^2$\citep{mukherjee:2020}, $^3$\citep{schroeder:2022}, $^4$\citep{gera:2022}, $^5$\citep{xiao:2023}. Column N represents the number of traning instances. 
}
\label{table-points-of-comparison}
\end{table}

Besides the final performance, it is also crucial to investigate the performance after each active learning iteration, which can be seen in the learning curves depicted in Figure~\ref{fig:learning-curves}. 
\stAcr{} self-training is a strong contender in most settings, especially with SetFit, where it sometimes reaches a performance close to the final value already during the first few iterations.
Besides the learning curves, a horizontal line at the top represents the model performance when training on the full train set, showing that several configurations achieve remarkable results at only 130 instances.
Moreover, the SetFit models are still slightly superior in this case, except for {TREC-6}, indicating that larger class imbalances could be a problem (at least for the hyperparameters that we used).
The corresponding area under curve values can be found in Appendix Table~\ref{table-self-training-results-auc}.
 
When comparing the results across query strategies, their impact seems to be minimal. The contrastive predictions strategy does not achieve superior performance in any configuration, so we focus on the comparison between breaking ties and random sampling. Breaking ties reaches slightly higher final scores at the expense of marginally lower area under curve. 

In Table~\ref{table-points-of-comparison}, we compare the best result per dataset to results from literature for sample-efficient methods, including zero-shot, few-shot, and active learning. 
Except for {TREC-6}, HAST achieves results close to the state of the art, despite using only very few instances and a comparably small model.
Notably, on AGN and DBP, HAST achieves results comparable to methods that used $525$ and $420$ instances, respectively, while using only 25\% and 30\% of those instances.  
HAST outperforms UST, AcTune, and VERIPS, while being on par with NeST. 
In combination with SetFit models, it even achieves slightly higher accuracy and F$_1$ scores. 

\section{Discussion}
\label{sec:discussion}

The experiments have shown that \stAcr{} is a highly effective self-training strategy that is able to leverage a large number of pseudo-labels.
When combined with SetFit models, \stAcr{} outperforms most other approaches and is on par with NeST in classification performance and area under curve.
Further investigation revealed that these gains can likely be attributed to the large number of pseudo-labels (as shown in Appendix Table~\ref{table-self-training-results-num-pseudo-labels}), whose acquisition is facilitated by the semantically meaningful embedding space. 
The large number of pseudo-labels, however, is likely to increase the level of label noise, but \stAcr{} demonstrates robustness against this, tolerating up to 20\% incorrect labels, particularly when paired with SetFit (see Appendix~\ref{appendix:impact-of-label-noise}).

Through an extensive reproduction, we have also investigated the relative strength of UST, AcTune, VERIPS, and NeST in the context of active learning for text classification.
The strongest contender is NeST, which is on par with HAST but does not outperform, despite using a computationally more expensive pseudo-label acquisition.
The primary impediment of previous approaches appears to be an overreliance on confidence thresholds, which are difficult to optimize in active learning scenarios.

\paragraph{Why did the experiments not incorporate the most recent large language models of 1B or more parameters?}

With a total runtime of \totalRuntime{}, the experiments are already computationally expensive---despite using models that are considered small by today's standards---and would be infeasible with larger model sizes. 
Moreover, research has demonstrated that smaller models can outperform larger ones~\citep{hsieh:2023} when properly fine-tuned or distilled. Therefore, we prioritize model efficiency in our active learning research, which ultimately aims to support real-world annotation where smaller models provide a more accessible and practical solution.

\paragraph{Why did the experiments use only a single self-training iteration?}

While increasing the number of self-training iterations {\em{may}} further increase the classification performance, this also runs the risk of degradation~\citep{gera:2022,xu:2023}. 
For this reason, by using only a single self-training iteration we minimize the risk of degradation, thereby using self-training not to replace but to complement active learning, and in favor of real-world settings at only little additional computational costs.

\paragraph{For which real-world use cases is the proposed method a good fit?} As previously stated, when self-training is combined with active learning it introduces another source of error.
The most favorable is the transductive setting~\cite{tong:2001,kottke:2023}, where the models do not necessarily need to generalize on future unseen data e.g., when active learning is used for labeling corpora in social sciences~\cite{romberg:2022} or biomedicine~\citep{nachtegael:2024}.
For datasets, whose class balance is heavily skewed, it might not be optimal yet, but this remains to be investigated in future research.

\paragraph{Will the reduced need in labeled data ultimately render active learning obsolete?}

Although our work has shown that strong models can be trained using very few samples, this was conducted on established benchmark datasets that are relatively small in size and less challenging compared to real-world datasets, which may have hundreds of multi-label classes, hierarchies, or highly skewed class distributions.
While simpler tasks, such as two-class sentiment analysis, might be solvable with zero-shot learning, more complex problems will still benefit from active learning. 
Should text classification become able to tackle the majority of problems through zero-shot or few-shot, active learning will remain valuable for refining class definitions by providing high-quality labels for instances where the current model exhibits high uncertainty.

\section{Conclusions}

In this work, we devise and evaluate \stAcr{}, a new self-training approach that is tailored to contrastive learning and aims to generate a large number of pseudo-labels to enhance the efficiency of contrastive training.
We reproduce four existing self-training approaches and evaluate all  approaches on the task of active learning for text classification.
Using only small language models of 110M parameters and 25\% of instances used in previous work, the proposed approach achieves results close to the state of the art on three out of four datasets.

\newpage
\section*{Limitations}

This study is a not a replication, but a reproduction with slight deviations that provide comparable conditions.
While this makes previous approaches comparable for the first time, this also introduces the risk of deviations or errors in the code.

While the overall approach has shown to be highly effective, for an active learning study, it is unfortunate that this seems to be largely caused by data-efficient models leveraging the additional pseudo-labels, and only to a minor degree by the instances selected by the query strategy.
Nevertheless, this was previously unknown and motivates further research on finding a query strategy that might be more beneficial for self-training.

Finally, the proposed approach is targeted at single-label classification. 
Our heuristic for hard-label decisions is not applicable to the multi-label settings and would require a different heuristic.

\section*{Ethical Considerations}

This work presents a method that reduces annotation efforts and could be used for good or bad---similar to most methods.
In either scenario, our method would help to reduce the annotation efforts, however, all of this could be achieved through extensive labeling efforts, without our method.

Moreover, since self-training relies on algorithmically assigned pseudo-labels, the obtained pseudo-label distribution is dependent on the unknown true distribution of the dataset, which could be biased towards certain classes.
In this case, self-training might not only be prone to error propagation, but also might propagate class biases.

\section*{Acknowledgments}

The authors acknowledge the financial support by the Federal Ministry of Education and Research of Germany and by S\"achsische Staatsministerium f\"ur Wissenschaft, Kultur und Tourismus in the programme Center of Excellence for AI-research ``Center for Scalable Data Analytics and Artificial Intelligence Dresden/Leipzig'', project identification number: ScaDS.AI.

We would like to thank the Webis Group and the Leipzig Corpora Collection for providing GPU resources.
We are especially grateful to the anonymous reviewers for their highly constructive and valuable feedback.

\bibliography{arr24-self-training-for-active-learning-lit}
\appendix

\newpage
\section*{Supplementary Material}

In the following, we provide details for reproduction (Sections~A--D), supplementary analyses (Section~E), and an extended discussion (Section~F).

\section{Environment}

The experiments were conducted using CUDA 11.2 and a single NVIDIA A100 GPU per run.
Experiment code is written in Python and executed in a Python 3.8 environment.
All experiment code has been published on Github: \href{https://github.com/chschroeder/self-training-for-sample-efficient-active-learning}{https://github.com/chschroeder/self-training-for-sample-efficient-active-learning}. 

\begin{figure*}[b]
\includegraphics[width=\textwidth]{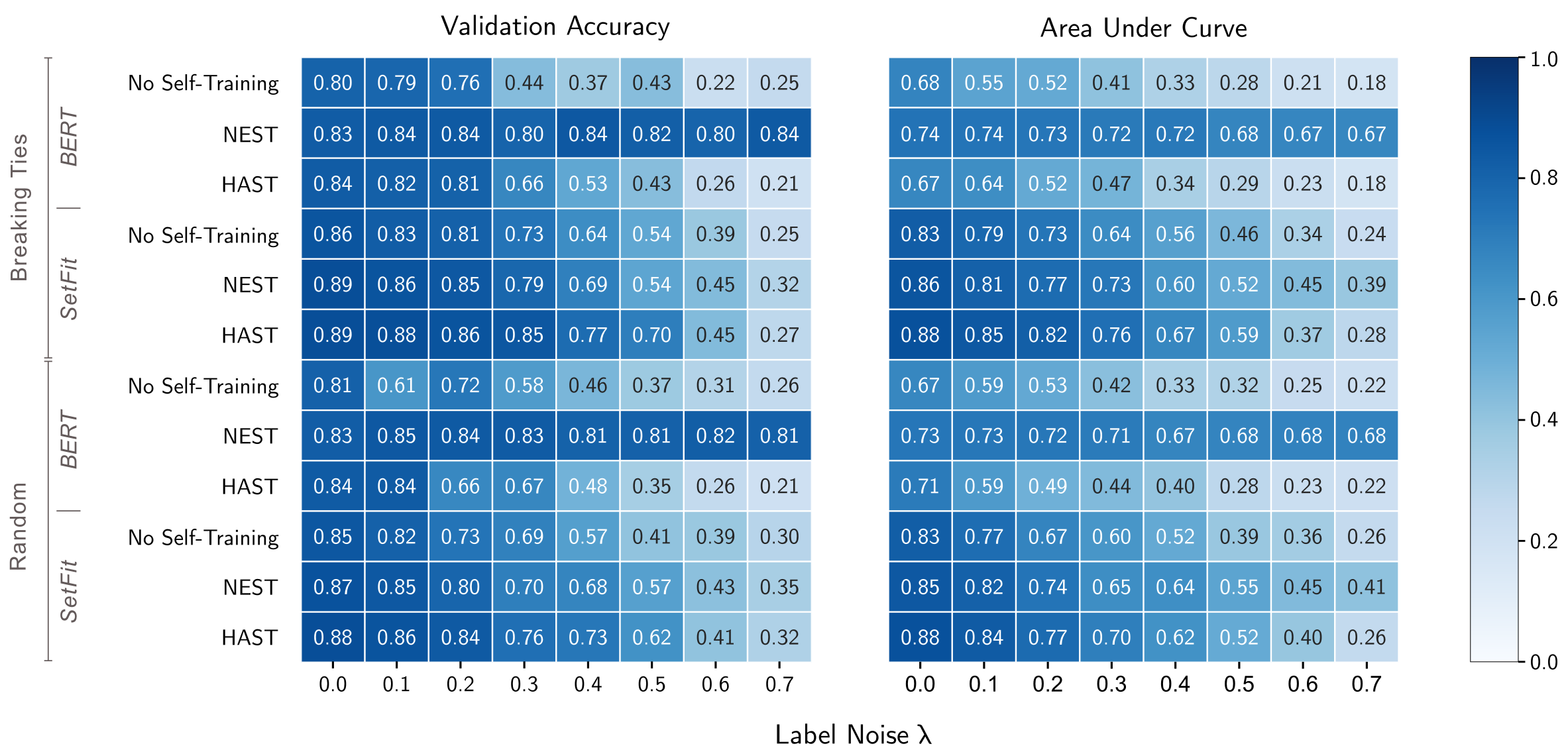}
\caption{The effect of label noise for NeST and HAST on AGN. Each label is replaced by an incorrect random label with probability $\lambda$. The left side shows validation accuracy after the final active learning iteration.
The right side shows the respective area under the learning curve for all 10 queries.}
\label{fig:label-noise}
\end{figure*}

\section{Software}

Our experiments leverage tried and test machine learning libraries: PyTorch~(2.2.1), transformers~(4.29.2), scikit-learn~(1.4.1.post1), setfit~(0.7.0), small-text~(2.0.0-dev, \href{https://github.com/webis-de/small-text/tree/f9be17a0d37eb630414dc92f2c528ffa52b09c15}{commit f9be17a0}), scipy~(1.12.0), numpy~(1.26.4). 
A full list, including transitive dependencies, is included in the Github repository.

The experiment code extends a previous code base~\cite{schroeder:2022}, which is built around the small-text library~\cite{schroeder:2023}. We extend this setup with self-training functionality, including the four reproduced strategies.

\section{Datasets}

The experiments used text classification benchmarks that are well-known and also widely used:  AG's News (AGN; \citealp{zhang:2015}), DBPedia~(DBP; \citealp{zhang:2015}), {IMDB}~\citep{maas:2011}, and {TREC-6}. \citep{li:2002}
The raw texts were obtained via the \href{https://github.com/huggingface/datasets}{huggingface datasets} library.
Following~\cite{margatina:2022}, we subsampled DBP to 10K instances per class (140K in total) to render the computational efforts (which are outlined in Section~\ref{appendix:experiments}) feasible.\\[1em]

\begin{table}[!htbp]%
\centering%
\fontsize{9pt}{10pt}\selectfont%
\renewcommand{\tabcolsep}{7pt}%
\begin{tabular}[t]{@{}l@{\hspace{16pt}}rr@{}}
\toprule
\bfseries Dataset & \bfseries Batch Size & \bfseries Max. Seq. Length\\
\midrule
AGN & 40 & 64 \\
DBP & 24 & 128   \\
IMDB & 14 & 512 \\
TREC & 40 & 64\\
    \bottomrule
\end{tabular}
\caption{Hyperparameter settings for the maximum sequence length (as number of tokens) per dataset.}
\label{table-sequence-length}
\end{table}

\section{Hyperparameters}

\paragraph{Self-Training}

For VERIPS, we used the margin-based variant, which has been shown to be superior to the entropy-based variant~\citep{gilhuber:2022}.

\paragraph{Maximum Sequence Length}

We set the maximum sequence length to the minimum multiple power of two for which 95\% of the dataset's sentences contain less than or an equal number of tokens, capped at 512 which is an architectural restriction of employed models~(see Table~\ref{table-sequence-length}).

\subsection{Classification}

\paragraph{BERT} 
We fine-tuned each model from \href{https://huggingface.co/google-bert/bert-base-uncased}{bert-base-uncased} using a learning rate of $\eta$~=~$\num{2e-5}$ and training the model for $15$~epochs.

\paragraph{SetFit} We follow the original publication~\cite{tunstall:2022} and first train the embeddings, which are subsequently used as features in combination with a logistic regression classification head. The original implementation was extended to support the per-instance weighting.

We use a learning rate of $\eta$~=~$\num{2e-5}$ and train for 1 epoch, which in the SetFit implementation is defined in iterations through the data. During each iteration, two pairs (one positive and one negative) are formed per labeled instance, which can lead to a steep increase in training data. For the sake of computational efficiency, we scale this parameter inversely to the number of pseudo-labels down to a minimum of $1$~iteration. 

\section{Experiments}
\label{appendix:experiments}

In Table~\ref{table-self-training-results-auc}, we provide the final area under curve values for the learning curves shown in Section~\ref{fig:learning-curves}.

\paragraph{Overall Runtime and GPU hours} The total runtime of all experiments configuration is \totalRuntime{}.

\paragraph{Individual Computational Costs} The average runtime of the training step (including self-training) is shown in Table~\ref{table-self-training-results-runtimes}.

\paragraph{Evaluation Metrics} We adhere to established active learning evaluation protocols and evaluate both the final classification performance and area under curve. The former is measured in accuracy for balanced datasets and in F$_1$ for imbalanced datasets.
For both metrics, tried and tested implementations from scikit-learn were used.

\subsection{Impact of Label Noise}
\label{appendix:impact-of-label-noise}

In the simulated active learning experiments, the annotation is realized by a simple lookup of the true labels. 
In real-world settings, however, answers provided by annotators may be wrong, either due to human label variation~\citep{plank:2022} or annotators making mistakes. 
Pseudo-labels are an imperfect heuristic, and especially in combination with self-training, those labels may be wrong---even disregarding human annotation errors, which may introduce additional noise.

For this reason, we investigate the effect of erroneous labeling in the annotation step and introduce a label noise $\lambda$, which represents the probability of a label to be wrong, i.e. replaced by random label other than the true label.
We investigate the two strongest self-training approaches  from Section~\ref{sec:experiments}: HAST and NeST. 
In Figure~\ref{fig:label-noise}, we present validation accuracy and AUC, broken down by increasing label noise. We find that up to a noise level of $\lambda=0.2$, HAST is only affected to smaller degree in AUC, while accuracy is only slightly lower. 
In the two rows where NeST is applied in combination with BERT, self-training fails since NeST is not able to find pseudo-labels, which is why the results are considerably better. 
This also shows the potential risk of self-training---especially when facing high label noise.

\subsection{Instance Weighting Ablation}

In Table~\ref{table-hast-ablations}, we present an ablation study over the weighting terms introduced in Section~\ref{sec:hard-label-regularization}.
Here we use HAST with the best performing query strategy, breaking ties, and ablate \Ni~class weights (by setting $\alpha=1$), \Nii~pseudo-label weights (by setting $\beta=1$), \Niii~class and pseudo-label weights (by setting $\alpha=\beta=1$).
Surprisingly, using both weightings simultaneously, does not yield the best results. 
Pseudo-label down-weighting seems to have more impact in general. 
Most importantly, it seems that weighting has a larger impact on BERT, while SetFit results are often close---except for the highly imbalanced dataset TREC.
Since, as reported in Section~\ref{sec:experiments}, even the fully supervised SetFit models seem to perform subpar on TREC-6, this is likely already a problem at the level of the classifier.

\section{Extended Discussion}

\paragraph{Why does HAST introduce a confidence threshold, despite the paper criticized previous methods for this?} At some point in every self-training algorithm, a decision on how to assign pseudo-labels is required.
We use a threshold of $s_i > 0.5$, the well-known tried and tested binary decision threshold, which is used in many classification settings.
Being the middle of the $[0, 1]$ confidence interval, this is the weakest decision criterion possible, and more importantly, it not optimized on the datasets (and not intended to).

\begin{table*}[!ht]
{
\centering%
\fontsize{9pt}{10pt}\selectfont%
\begin{tabular}[h]{@{}ll@{\hspace{3mm}}rrrr@{}}
\toprule
& & \multicolumn{4}{@{}c@{}}{\bfseries Datasets}\\
\cmidrule{3-6}\bfseries Classifier & \bfseries Self-Training & \bfseries AGN & \bfseries DBP & \bfseries IMDB & \bfseries TREC\\
\midrule
\multicolumn{6}{@{}c@{}}{\bfseries Final Accuracy/F$_1$}\\
\midrule
\multirow{4}{*}{BERT} & HAST & 0.766\stddev{0.063} & 0.604\stddev{0.067} & 0.807\stddev{0.109} & 0.405\stddev{0.192}\\
 & ~w/o class weighting ($\alpha = 1.0$) & 0.845\stddev{0.008} & \bfseries 0.794\stddev{0.048} & 0.799\stddev{0.097} & 0.589\stddev{0.160}\\
 & ~w/o pseudo-label down-weighting ($\beta = 1.0$) & 0.855\stddev{0.017} & \bfseries 0.794\stddev{0.048} & 0.695\stddev{0.139} & \bfseries 0.601\stddev{0.112}\\
 & ~w/o class weighting and down-weighting ($\alpha = \beta = 1.0$) & \bfseries 0.859\stddev{0.019} & \bfseries 0.794\stddev{0.048} & \bfseries 0.849\stddev{0.016} & 0.536\stddev{0.091}\\
 \cmidrule{2-6} \multirow{4}{*}{SetFit} & HAST & 0.889\stddev{0.006} & 0.984\stddev{0.001} & 0.881\stddev{0.045} & 0.691\stddev{0.012}\\
 & ~w/o class weighting ($\alpha = 1.0$) & 0.886\stddev{0.003} & \bfseries 0.985\stddev{0.003} & \bfseries 0.924\stddev{0.004} & 0.763\stddev{0.015}\\
 & ~w/o pseudo-label down-weighting ($\beta = 1.0$) & \bfseries 0.889\stddev{0.002} & 0.983\stddev{0.004} & 0.914\stddev{0.009} & 0.761\stddev{0.009}\\
 & ~w/o class weighting and down-weighting ($\alpha = \beta = 1.0$) & 0.889\stddev{0.002} & 0.985\stddev{0.001} & 0.902\stddev{0.031} & \bfseries 0.785\stddev{0.019}\\
\midrule
\multicolumn{6}{@{}c@{}}{\bfseries Area under Curve}\\
\midrule
\multirow{4}{*}{BERT} & HAST & 0.634\stddev{0.034} & 0.332\stddev{0.018} & 0.670\stddev{0.037} & 0.278\stddev{0.029}\\
 & ~w/o class weighting ($\alpha = 1.0$) & 0.683\stddev{0.012} & \bfseries 0.484\stddev{0.037} & \bfseries 0.711\stddev{0.032} & \bfseries 0.393\stddev{0.046}\\
 & ~w/o pseudo-label down-weighting ($\beta = 1.0$) & 0.651\stddev{0.042} & \bfseries 0.484\stddev{0.037} & 0.690\stddev{0.029} & 0.381\stddev{0.041}\\
 & ~w/o class weighting and down-weighting ($\alpha = \beta = 1.0$) & \bfseries 0.691\stddev{0.009} & \bfseries 0.484\stddev{0.037} & 0.700\stddev{0.026} & 0.392\stddev{0.036}\\
 \cmidrule{2-6} \multirow{4}{*}{SetFit} & HAST & \bfseries 0.873\stddev{0.004} & 0.942\stddev{0.015} & \bfseries 0.898\stddev{0.004} & 0.636\stddev{0.023}\\
 & ~w/o class weighting ($\alpha = 1.0$) & 0.871\stddev{0.005} & \bfseries 0.951\stddev{0.009} & 0.891\stddev{0.008} & \bfseries 0.737\stddev{0.012}\\
 & ~w/o pseudo-label down-weighting  ($\beta = 1.0$) & 0.870\stddev{0.003} & 0.937\stddev{0.017} & 0.897\stddev{0.008} & 0.714\stddev{0.014}\\
 & ~w/o class weighting and down-weighting  ($\alpha = \beta = 1.0$) & 0.869\stddev{0.005} & 0.940\stddev{0.009} & 0.897\stddev{0.008} & 0.721\stddev{0.020}\\
\bottomrule
\end{tabular}
\caption{Ablation analysis: final classification performance (top) in accuracy or macro-F$_1$ and area under curve (bottom) when removing different components from the instance weighting (see Section~\ref{sec:hard-label-regularization}). Breaking ties was employed as query strategy for all runs and the reported numbers are the average over five runs. The reported numbers represent the average over five runs, with the standard deviations shown to the right of each value.}
\label{table-hast-ablations}
}
\end{table*}
\begin{table*}[!ht]
{
\centering%
\fontsize{9pt}{10pt}\selectfont%
\renewcommand{\tabcolsep}{10pt}%
\begin{tabular}[H]{@{}l@{\hspace{11mm}}l@{\hspace{12mm}}r@{\hspace{8mm}}rrrr@{}}
\toprule
& & & \multicolumn{4}{c}{\bfseries Datasets}\\
\cmidrule{4-7}\bfseries Strategy & \bfseries Classifier & \bfseries Self-Training & \bfseries AGN & \bfseries DBP & \bfseries IMDB & \bfseries TREC\\\midrule
\multirow{12}{*}{Breaking Ties} & \multirow{6}{*}{BERT} & No Self-Training & 0.638\stddev{0.030} & 0.335\stddev{0.021} & 0.650\stddev{0.019} & 0.285\stddev{0.032}\\
 & & UST & 0.664\stddev{0.017} & 0.283\stddev{0.030} & 0.684\stddev{0.009} & 0.231\stddev{0.077}\\
 & & AcTune & 0.656\stddev{0.028} & 0.324\stddev{0.015} & 0.676\stddev{0.030} & 0.268\stddev{0.033}\\
 & & VERIPS & 0.728\stddev{0.019} & \bfseries 0.640\stddev{0.026} & \bfseries 0.701\stddev{0.010} & 0.465\stddev{0.063}\\
 & & NeST & \bfseries 0.733\stddev{0.011} & 0.638\stddev{0.047} & 0.695\stddev{0.025} & \bfseries 0.467\stddev{0.028}\\
 & & HAST & 0.634\stddev{0.030} & 0.333\stddev{0.018} & 0.668\stddev{0.033} & 0.304\stddev{0.032}\\
 \cmidrule(){3-7} & \multirow{6}{*}{SetFit} & No Self-Training & 0.818\stddev{0.009} & 0.830\stddev{0.016} & 0.868\stddev{0.003} & 0.628\stddev{0.021}\\
 & & UST & 0.573\stddev{0.006} & 0.295\stddev{0.012} & 0.794\stddev{0.008} & 0.394\stddev{0.020}\\
 & & AcTune & 0.831\stddev{0.006} & 0.906\stddev{0.005} & 0.886\stddev{0.007} & 0.548\stddev{0.021}\\
 & & VERIPS & 0.825\stddev{0.006} & 0.851\stddev{0.022} & 0.877\stddev{0.008} & 0.650\stddev{0.017}\\
 & & NeST & 0.840\stddev{0.006} & 0.865\stddev{0.016} & \bfseries 0.917\stddev{0.002} & \bfseries 0.728\stddev{0.031}\\
 & & HAST & \bfseries 0.871\stddev{0.002} & \bfseries 0.942\stddev{0.015} & 0.898\stddev{0.004} & 0.711\stddev{0.010}\\
 \midrule
 \multirow{12}{*}{\parbox{1.8cm}{Contrastive Predictions}} & \multirow{6}{*}{BERT} & No Self-Training & 0.498\stddev{0.036} & 0.228\stddev{0.037} & 0.640\stddev{0.021} & 0.203\stddev{0.026}\\
  & & UST & 0.594\stddev{0.041} & 0.148\stddev{0.035} & 0.665\stddev{0.014} & 0.195\stddev{0.046}\\
  & & AcTune & 0.550\stddev{0.043} & 0.222\stddev{0.032} & 0.642\stddev{0.015} & 0.208\stddev{0.026}\\
  & & Verips & \bfseries 0.678\stddev{0.021} & \bfseries 0.451\stddev{0.042} & \bfseries 0.676\stddev{0.026} & 0.366\stddev{0.058}\\
  & & NeST & 0.598\stddev{0.060} & 0.424\stddev{0.037} & 0.661\stddev{0.023} & \bfseries 0.386\stddev{0.073}\\
  & & HAST & 0.566\stddev{0.052} & 0.232\stddev{0.021} & 0.675\stddev{0.023} & 0.233\stddev{0.082}\\
  \cmidrule(){3-7}  & \multirow{6}{*}{SetFit} & No Self-Training & 0.757\stddev{0.003} & 0.643\stddev{0.014} & 0.849\stddev{0.009} & 0.573\stddev{0.019}\\
  & & UST & 0.537\stddev{0.017} & 0.250\stddev{0.014} & 0.769\stddev{0.018} & 0.345\stddev{0.031}\\
  & & AcTune & 0.785\stddev{0.011} & 0.623\stddev{0.020} & 0.877\stddev{0.010} & 0.608\stddev{0.046}\\
  & & Verips & 0.765\stddev{0.009} & 0.708\stddev{0.022} & 0.880\stddev{0.011} & 0.621\stddev{0.026}\\
  & & NeST & 0.780\stddev{0.002} & 0.722\stddev{0.036} & \bfseries 0.907\stddev{0.004} & 0.622\stddev{0.017}\\
  & & HAST & \bfseries 0.845\stddev{0.007} & \bfseries 0.814\stddev{0.046} & 0.904\stddev{0.005} & \bfseries 0.714\stddev{0.016}\\
 \midrule 
 \multicolumn{7}{c}{ (Continued on next page.)}\\
\end{tabular}
}
\end{table*}

\begin{table*}[!ht]
{
\centering%
\fontsize{9pt}{10pt}\selectfont%
\renewcommand{\tabcolsep}{10pt}%
\begin{tabular}[H]{@{}l@{\hspace{15mm}}l@{\hspace{15mm}}r@{\hspace{8mm}}rrrr@{}}
\multicolumn{7}{c}{ (Continued from previous page.)}\\
\toprule
& & & \multicolumn{4}{c}{\bfseries Datasets}\\
\cmidrule{4-7}\bfseries Strategy & \bfseries Classifier & \bfseries Self-Training & \bfseries AGN & \bfseries DBP & \bfseries IMDB & \bfseries TREC\\\midrule
\multirow{12}{*}{Random} & \multirow{6}{*}{BERT} & No Self-Training & 0.630\stddev{0.013} & 0.307\stddev{0.028} & 0.653\stddev{0.024} & 0.263\stddev{0.038}\\
    & & UST & 0.681\stddev{0.030} & 0.330\stddev{0.034} & 0.694\stddev{0.010} & 0.233\stddev{0.021}\\
    & & AcTune & 0.648\stddev{0.023} & 0.335\stddev{0.027} & 0.658\stddev{0.026} & 0.284\stddev{0.031}\\
    & & VERIPS & 0.727\stddev{0.013} & \bfseries 0.608\stddev{0.051} & \bfseries 0.697\stddev{0.033} & \bfseries 0.446\stddev{0.049}\\
    & & NeST & \bfseries 0.735\stddev{0.022} & 0.592\stddev{0.034} & 0.676\stddev{0.024} & 0.428\stddev{0.056}\\
    & & HAST & 0.625\stddev{0.042} & 0.354\stddev{0.027} & 0.679\stddev{0.039} & 0.301\stddev{0.041}\\
    \cmidrule(){3-7} & \multirow{6}{*}{SetFit} & No Self-Training & 0.814\stddev{0.009} & 0.755\stddev{0.038} & 0.885\stddev{0.006} & 0.608\stddev{0.023}\\
    & & UST & 0.577\stddev{0.008} & 0.285\stddev{0.020} & 0.811\stddev{0.020} & 0.381\stddev{0.026}\\
    & & AcTune & 0.820\stddev{0.012} & 0.774\stddev{0.028} & 0.912\stddev{0.001} & 0.547\stddev{0.018}\\
    & & VERIPS & 0.823\stddev{0.007} & 0.916\stddev{0.015} & 0.912\stddev{0.002} & 0.649\stddev{0.021}\\
    & & NeST & 0.834\stddev{0.007} & 0.916\stddev{0.012} & \bfseries 0.914\stddev{0.005} & \bfseries  0.763\stddev{0.008}\\
    & & HAST & \bfseries 0.868\stddev{0.006} & \bfseries 0.941\stddev{0.010} & 0.911\stddev{0.007} & 0.693\stddev{0.021}\\
\bottomrule
\end{tabular}
\caption{Area under curve per query strategy, classifier, self-training method, and dataset. For AGN and IMDB the area under the accuracy curve is listed, for DBP and TREC the area under the macro-F$_1$ curve. The reported numbers represent the average over five runs, with the standard deviations shown to the right of each value.}
\label{table-self-training-results-auc}
}
\vspace*{0.5cm}
\end{table*}

\begin{table*}[!hptb]
{
\centering%
\fontsize{9pt}{10pt}\selectfont%
\renewcommand{\tabcolsep}{10pt}%
\begin{tabular}[H]{@{}l@{\hspace{8mm}}l@{\hspace{12mm}}rrrrr@{}}
  \toprule
  & & & \multicolumn{4}{c}{\bfseries Datasets}\\
  \cmidrule{4-7}\bfseries Strategy & \bfseries Classifier & \bfseries Self-Training & \bfseries AGN & \bfseries DBP & \bfseries IMDB & \bfseries TREC\\
  \midrule
  \multirow{10}{*}{Breaking Ties} & \multirow{5}{*}{BERT} & UST & 352.24\stddev{1.11} & 970.27\stddev{3.81} & 1879.50\stddev{3.14} & 98.25\stddev{1.36}\\
   & & AcTune & 301.39\stddev{13.73} & 692.87\stddev{5.08} & 1013.13\stddev{42.33} & 36.61\stddev{0.83}\\
   & & VERIPS & 252.08\stddev{2.94} & 723.46\stddev{3.57} & 837.57\stddev{1.79} & 62.17\stddev{2.13}\\
   & & NeST & 244.58\stddev{1.86} & 723.03\stddev{3.26} & 846.71\stddev{4.91} & 63.70\stddev{0.99}\\
   & & HAST & 240.71\stddev{2.12} & 690.35\stddev{3.71} & 1897.03\stddev{89.95} & 41.48\stddev{3.48}\\
   \cmidrule(){3-7} & \multirow{5}{*}{SetFit} &  UST & 716.14\stddev{2.34} & 1312.27\stddev{3.68} & 2655.04\stddev{7.94} & 42.71\stddev{0.24}\\
   & & AcTune & 673.86\stddev{1.07} & 1475.65\stddev{2.91} & 1578.38\stddev{4.40} & 17.68\stddev{0.48}\\
   & & VERIPS & 453.75\stddev{1.66} & 1021.13\stddev{2.43} & 1238.13\stddev{2.12} & 13.67\stddev{0.14}\\
   & & NeST & 465.65\stddev{2.61} & 1011.67\stddev{3.52} & 1242.64\stddev{3.68} & 16.03\stddev{0.30}\\
   & & HAST & 677.57\stddev{5.97} & 1123.51\stddev{3.71} & 2579.80\stddev{10.94} & 25.27\stddev{0.33}\\
  \midrule
  \multirow{10}{*}{\parbox{1.8cm}{Contrastive Predictions}} & \multirow{6}{*}{BERT} & UST & 353.84\stddev{4.89} & 948.33\stddev{7.66} & 1882.40\stddev{5.53} & 97.87\stddev{3.00}\\
 & & AcTune & 353.40\stddev{3.46} & 691.46\stddev{1.41} & 1059.80\stddev{8.37} & 37.00\stddev{0.74}\\
 & & Verips & 245.26\stddev{1.14} & 730.82\stddev{3.11} & 834.88\stddev{4.35} & 64.69\stddev{0.58}\\
 & & NeST & 250.04\stddev{2.00} & 739.44\stddev{1.82} & 841.60\stddev{5.00} & 64.84\stddev{0.59}\\
 & & HAST & 244.35\stddev{1.78} & 694.14\stddev{1.68} & 1967.57\stddev{46.02} & 39.69\stddev{0.44}\\
 \cmidrule(){3-7} & \multirow{5}{*}{SetFit} & UST & 710.34\stddev{4.07} & 1290.36\stddev{1.92} & 2653.14\stddev{15.72} & 41.31\stddev{0.75}\\
 & & AcTune & 667.61\stddev{1.65} & 1463.93\stddev{1.88} & 1564.77\stddev{5.86} & 15.07\stddev{0.11}\\
 & & Verips & 454.07\stddev{1.20} & 994.89\stddev{1.71} & 1238.64\stddev{6.81} & 13.20\stddev{0.16}\\
 & & NeST & 471.62\stddev{1.01} & 1010.48\stddev{3.86} & 1236.69\stddev{6.40} & 14.95\stddev{0.18}\\
 & & HAST & 660.91\stddev{5.54} & 1040.48\stddev{8.59} & 2577.87\stddev{12.03} & 23.56\stddev{0.63}\\
\midrule
\multirow{10}{*}{Random} & \multirow{5}{*}{BERT} & UST & 350.81\stddev{2.66} & 947.98\stddev{1.36} & 1881.25\stddev{3.22} & 96.76\stddev{2.98}\\
  & & AcTune & 320.59\stddev{18.36} & 696.55\stddev{4.64} & 1015.37\stddev{31.33} & 36.29\stddev{1.22}\\
  & & VERIPS & 261.08\stddev{5.14} & 726.11\stddev{5.94} & 836.54\stddev{2.41} & 94.99\stddev{5.26}\\
  & & NeST & 265.89\stddev{5.44} & 726.38\stddev{4.50} & 847.49\stddev{4.60} & 62.09\stddev{0.98}\\
  & & HAST & 240.68\stddev{2.14} & 691.40\stddev{0.62} & 1840.03\stddev{62.39} & 42.86\stddev{3.26}\\
  \cmidrule(){3-7} & \multirow{5}{*}{SetFit} & UST & 720.45\stddev{5.89} & 1304.55\stddev{1.78} & 2663.02\stddev{4.80} & 40.90\stddev{0.76}\\
  & & AcTune & 685.87\stddev{6.17} & 1492.17\stddev{2.82} & 1573.35\stddev{1.39} & 18.43\stddev{0.28}\\
  & & VERIPS & 457.66\stddev{0.43} & 993.81\stddev{3.14} & 1234.64\stddev{2.01} & 13.15\stddev{0.07}\\
  & & NeST & 471.96\stddev{1.48} & 1011.17\stddev{2.69} & 1357.29\stddev{8.12} & 15.29\stddev{0.32}\\
  & & HAST & 703.01\stddev{7.19} & 1150.31\stddev{7.58} & 2676.58\stddev{14.91} & 27.17\stddev{0.46}\\
\bottomrule
\end{tabular}
\caption{Mean average training runtime (in seconds) over all iterations. A failed effort to obtain pseudo-labels is counted as zero seconds and therefore reduces the runtime. The reported numbers represent the average over five runs, with the standard deviations shown to the right of each value.}
\label{table-self-training-results-runtimes} 
}
\end{table*}

\begin{table*}[t]
{
\centering%
\fontsize{9pt}{10pt}\selectfont%
\renewcommand{\tabcolsep}{10pt}%
\begin{tabular}[t]{@{}l@{\hspace{4mm}}l@{\hspace{5mm}}rrrrr@{}}
  \toprule
  & & & \multicolumn{4}{c}{\bfseries Datasets}\\
  \cmidrule{4-7}\bfseries Strategy & \bfseries Classifier & \bfseries Self-Training & \bfseries AGN & \bfseries DBP & \bfseries IMDB & \bfseries TREC\\
  \midrule
  \multirow{10}{*}{Breaking Ties} & \multirow{5}{*}{BERT} & UST & 14.82\stddev{38.19} & 215.92\stddev{127.58} & 6.50\stddev{25.40} & 168.96\stddev{45.49}\\
   & & AcTune & 5.23\stddev{11.84} & 0.00\stddev{0.00} & 4.40\stddev{13.32} & 0.00\stddev{0.00}\\
   & & VERIPS & 0.00\stddev{0.00} & 0.00\stddev{0.00} & 0.00\stddev{0.00} & 0.00\stddev{0.00}\\
   & & NeST & 0.00\stddev{0.00} & 0.00\stddev{0.00} & 6.96\stddev{6.91} & 0.00\stddev{0.00}\\
   & & HAST & 620.80\stddev{1136.78} & 0.00\stddev{0.00} & 1172.96\stddev{4169.12} & 138.00\stddev{167.35}\\
   \cmidrule(){3-7}  & \multirow{5}{*}{SetFit} & UST & 1.06\stddev{0.04} & 24.77\stddev{29.16} & 1.08\stddev{0.06} & 61.04\stddev{6.81}\\
   & & AcTune & 1.72\stddev{0.34} & 2.89\stddev{0.64} & 1.20\stddev{0.20} & 7.20\stddev{4.95}\\
   & & VERIPS & 0.00\stddev{0.00} & 0.00\stddev{0.00} & 0.00\stddev{0.00} & 0.00\stddev{0.00}\\
   & & NeST & 3.15\stddev{1.47} & 3.76\stddev{1.39} & 2.03\stddev{1.06} & 7.36\stddev{4.19}\\
   & & HAST & 1.49\stddev{0.45} & 90.91\stddev{116.36} & 1.14\stddev{0.15} & 292.69\stddev{222.55}\\
   \midrule
   \multirow{10}{*}{\parbox{1.8cm}{Contrastive Predictions}} & \multirow{5}{*}{BERT} & UST & 120.0\stddev{0.0} & 420.0\stddev{0.0} & 60.0\stddev{0.0} & 180.0\stddev{0.0}\\
    & & AcTune & 24.9\stddev{0.2} & 0.0\stddev{0.0} & 25.0\stddev{0.0} & 0.0\stddev{0.0}\\
    & & Verips & 0.0\stddev{0.0} & 0.0\stddev{0.0} & 0.0\stddev{0.0} & 0.0\stddev{0.0}\\
    & & NeST & 0.0\stddev{0.0} & 0.0\stddev{0.0} & 536.5\stddev{443.3} & 0.0\stddev{0.0}\\
    & & HAST & 1391.0\stddev{1085.6} & 0.0\stddev{0.0} & 13741.9\stddev{1268.1} & 90.0\stddev{116.8}\\
    \cmidrule(){3-7} & \multirow{5}{*}{SetFit} & UST & 120.0\stddev{0.0} & 420.0\stddev{0.0} & 60.0\stddev{0.0} & 180.0\stddev{0.0}\\
    & & AcTune & 25.0\stddev{0.0} & 25.0\stddev{0.3} & 25.0\stddev{0.0} & 25.0\stddev{0.0}\\
    & & Verips & 0.0\stddev{0.0} & 0.0\stddev{0.0} & 0.0\stddev{0.0} & 0.0\stddev{0.0}\\
    & & NeST & 81.6\stddev{78.3} & 19.8\stddev{22.7} & 155.9\stddev{102.2} & 22.8\stddev{14.0}\\
    & & HAST & 9178.5\stddev{1000.0} & 1964.9\stddev{1312.3} & 14355.7\stddev{306.4} & 1507.5\stddev{337.0}\\
  \midrule
  \multirow{10}{*}{Random} & \multirow{5}{*}{BERT} & UST & 12.13\stddev{33.69} & 164.45\stddev{106.32} & 5.84\stddev{22.03} & 165.05\stddev{46.27}\\
   & & AcTune & 2.90\stddev{4.07} & 0.00\stddev{0.00} & 2.54\stddev{7.91} & 0.00\stddev{0.00}\\
   & & VERIPS & 0.00\stddev{0.00} & 0.00\stddev{0.00} & 1.05\stddev{0.08} & 0.00\stddev{0.00}\\
   & & NeST & 0.00\stddev{0.00} & 0.00\stddev{0.00} & 6.72\stddev{6.12} & 0.00\stddev{0.00}\\
   & & HAST & 469.14\stddev{750.86} & 0.00\stddev{0.00} & 2105.59\stddev{5268.35} & 72.53\stddev{79.30}\\
   \cmidrule(){3-7}  & \multirow{5}{*}{SetFit} & UST & 1.06\stddev{0.04} & 23.12\stddev{30.56} & 1.09\stddev{0.06} & 61.51\stddev{6.92}\\
   & & AcTune & 1.54\stddev{0.30} & 4.86\stddev{2.59} & 1.23\stddev{0.25} & 10.05\stddev{7.46}\\
   & & VERIPS & 0.00\stddev{0.00} & 0.00\stddev{0.00} & 1.18\stddev{0.00} & 0.00\stddev{0.00}\\
   & & NeST & 5.07\stddev{3.22} & 13.25\stddev{10.74} & 3.69\stddev{3.06} & 9.46\stddev{5.95}\\
   & & HAST & 1.63\stddev{0.40} & 86.84\stddev{142.18} & 1.07\stddev{0.06} & 333.54\stddev{255.17}\\
  \bottomrule
  \end{tabular}
\caption{Mean average number of pseudo labels over all iterations, broken down per query strategy, classifier, and self-training approach. A value of zero indicates that no pseudo-labels could be selected, mostly due to not exceeding the confidence threshold. The reported numbers represent the average over five runs, with the standard deviations shown to the right of each value.}
\label{table-self-training-results-num-pseudo-labels}
}
\vspace*{9.8cm}
\end{table*}

\end{document}